# Estimating Countries with Similar Maternal Mortality Rate using Cluster Analysis and Pairing Countries with Identical MMR

Dr. S. Nandini

Associate Professor in Zoology

Quaid-E-Millath Government College for Women, Anna Salai, Chennai - 600 002

s.nandini0507@gmail.com

Sanjjushri Varshini R

sanjjushrivarshini@gmail.com

**Abstract:**

In the evolving world, we require more additionally the young era to flourish and evolve into developed land. Most of the population all around the world are unaware of the complications involved in the routine they follow while they are pregnant and how hospital facilities affect maternal health. Maternal Mortality is the death of a pregnant woman due to intricacies correlated to pregnancy, underlying circumstances exacerbated by the pregnancy or management of these situations. It is crucial to consider the Maternal Mortality Rate (MMR) in diverse locations and determine which human routines and hospital facilities diminish the Maternal Mortality Rate (MMR). This research aims to examine and discover the countries which are keeping more lavish threats of MMR and countries alike in MMR encountered. Data is examined and collected for various countries, data consists of the earlier years' observation. From the perspective of Machine Learning, Unsupervised Machine Learning is implemented to perform Cluster Analysis. Therefore the pairs of countries with similar MMR as well as the extreme opposite pair concerning the MMR are found.

**1 Introduction:**

A pregnant woman's death will affect her and the entire world in some means. The cause of maternal death is ambiguous for most of the course. It occasionally even becomes challenging for doctors to determine the reason for the cause behind the death. Over the past years, the Maternal Mortality Rate (MMR) has decreased in some provinces whereas the same has risen in other provinces. It is not just one or two factors that affect MMR, and thus when the comparison is made among the region's establishments and routines the individuals who live there undergo to estimate which plays a major role in reducing maternal death by a massive count.

The method of figuring out the countries which fall under the same category and vice versa will the government and the women who are pregnant so that precautions can be taken to reduce maternal death.

The integration of Data Science to solve medical problems has especially helped in several ways. It is significant to accomplish an examination of the Maternal Mortality Rate over the year and around the world so that we will be able to find countries that will have to adapt to what type of hospitality features. To find these unsupervised machine learning is implemented in an advanced form i.e., by using different clustering methods.

Still, when it comes to the analysis of MMR country-wise, it's identified that there's been an incremental increase in MMR in many countries whereas there's a decline in several countries too. This research is based on finding the causes of this gradient in countries around the World. There are many grounds why the drop in MMR takes place. These justifications are to be interpreted in this study by analysing and pairing countries having a significant change and the opposite MMR rate in the pairing of countries is found. Moreover, the main causes of the decline in Maternal Mortality Rate include the lack of hospital care, infections and complications during delivery, or even unattended abortions. These causes might differ according to different locations.

To address these issues this research is implemented effectively. The data is gathered from the World Health Organisation (WHO) from the year 1990 in 61 countries.

Even though each country has their MMR per year, when we compute results from the different years and compare them with the other countries we can effectively conclude the factors which hospitality facilities, environmental conditions, women's food habits and mental habits are better or poor and how it affects the pregnant women. This will proclaim the unrecognised facts for both the public as well as the government and doctors to know where things lack.

**2 Literature Review:**

[1]A study to prevent pregnancy disorders to reduce the risk of maternal and infant mortality all over the world is done. Preeclampsia and other hypertensive pregnancy disorders have been explicitly examined. The truth behind this disorder will help to reduce the maternal mortality rate by developing solutions to the disorder. [2] The analysis and prediction of the Maternal Mortality Rate (MMR) in India are performed using the ensemble (Machine Learning framework) that combines various algorithms for effective performance and exact estimation. The main aim here was to estimate districts with high and low MMR over the Indian Region. [3] By developing

methodology using vital statistics maternal mortality data considering the changes in pregnancy question formats over time and between states. In their observational study, they analyzed vital statistics maternal mortality data from all U.S. states according to the format and year of adoption of the pregnancy question. [4] Nowadays, there are maternal problems due to health issues in pregnant women. These issues may arise due to complications either during pregnancy or delivery time or after delivery according to the women's health condition. This study uses different methodologies to improve maternal and fetal health irrespective of the place. The Indian data comprising MMR is used. Concerning the complications and symptoms, treatment is suggested accordingly. [5] Across the globe, about 830 women die per day due to pregnancy-related issues. This is with more than two-thirds of these cases occurring in Africa. Descriptive analysis is performed, and tables and maps are presented for the data of 54 African countries between 2006 and 2018. The results show that the average prevalence of Maternal Mortality Rate (MMR) in Africa was 415 per 100000 live births. Sustainable development is required concerning maternal health. [6] MMR is an important indicator used to govern maternal health. The government of Indonesia has made several efforts to pull down the MMR. Their study aimed to group the provinces in Indonesia based on maternal health indicators using Variable Weighting K-Means and Fuzzy C-Means. This result comprises five groups in total that indicate very low, low, medium, high, and very high scores concerning the maternal health indicator. [7] A pregnancy becomes successful when the biological adaptations are proper with the balance in the hormone level and immune. By 2030, the Sustainable Development Goal 3 (SDG 3) of the United Nations seeks to increase maternal health and decrease child and maternal mortality. Anyhow, maternal mortality has not decreased drastically, mainly in developing countries. They aimed to examine and deliver a method for monitoring and estimating the risk involved in the life of pregnant women. [8] In their study, they have taken into consideration reducing maternal mortality in the United States, more than 700 maternal deaths occur due to the intricacies and many women experience life-threatening complications. African women undergo more complications than white women, and this has not changed over a decade. [9] The majority of reason cause maternal mortality is found by tracking, socio-economic status, race, and hospital facilities that specifies the geological location. Machine Learning techniques are employed to identify the at-risk population previously before the complications of health take place. Binary classification is effectively used for validating examples and operates these classifications to predict maternal mortality risk. [10] Data Mining is one of the crucial sub-processes of the Knowledge Discovery Process methodology that is implemented for the algorithms on the target data. A methodology for extracting knowledge from the Maternal Health Dataset is proposed. The dataset used here is Indian data concerning maternal health. Machine Learning metrics like accuracy, F-measure, AUC, and Gini are used. [11] The crisis met by Indonesia is that the Maternal Mortality Rate (MMR) is higher than in other neighbouring countries. Maternal mortality indisputably illustrates the complexity associated with pregnancy. Maternal mortality has become a factor in the development of a country. If we reduce maternal mortality then the overall health status will improve. They aimed to classify maternal mortality health and consequently reduce the MMR in Dairi. Two supervised machine learning model decision tree and Naive Bayes is utilised. [12] About 99% of maternal deaths from pregnancy and childbirth-related avoidable causes take place in underdeveloped nations. Afghanistan has a high MMR. There exist several reasons for maternal death. Their study aims to provide a strategy to forecast the risk level of maternal complexity with the benefit of a machine learning methodology.

**3 Implementation:**

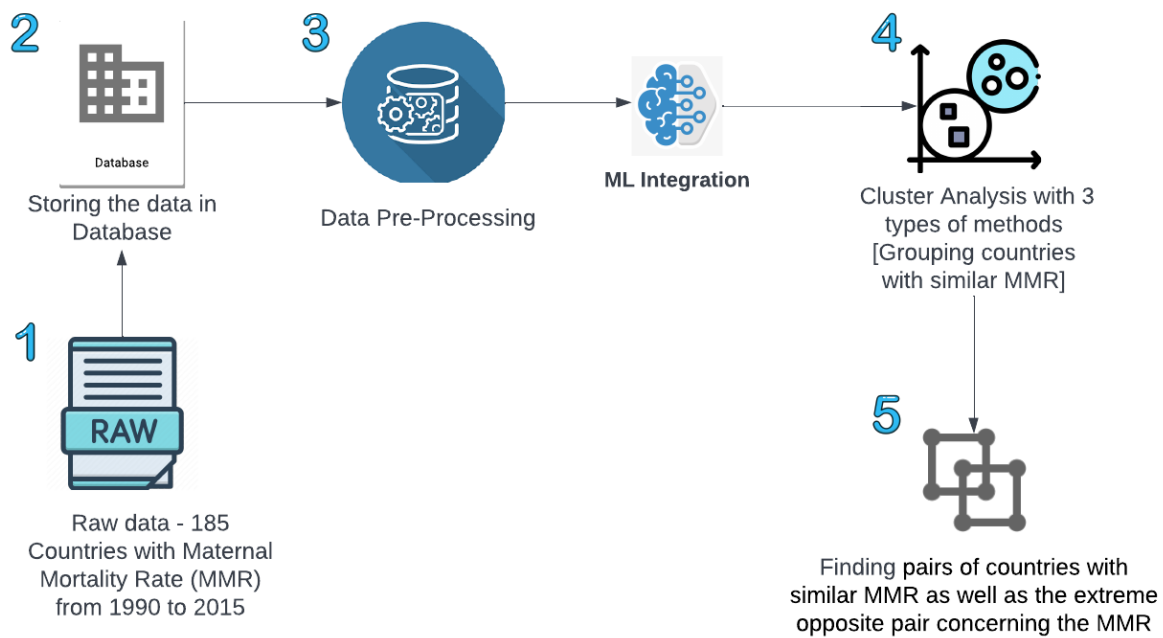

Figure 1: Architecture Diagram

The data is collected from the World Health Organisation (WHO) which comprises Maternal Mortality Rates from the year 1990 - 2015, in 185 countries around the world. The data is in an unclean format.

The collected data is stored in the PostgreSQL database. Explicitly PostgreSQL is chosen for advanced manipulation operation inbuilt feature. All the pre-processing steps for cleaning the data so that the data can be fed to the Machine Learning Model directly for further development. The task of arranging a set of data into the same group so that they are more comparable to one another than to those in different groups is known as cluster analysis or clustering. It is a key pursuit of exploratory data analysis and a typical statistical data analysis technique. Mainly three types of clustering methods are performed to bring out the best model. The countries with similar Maternal Mortality rates over the years are grouped. We are Finding pairs of countries with similar Maternal Mortality Rates as well as countries with extremely opposite pairs concerning the Maternal Mortality Rate.

**4 Models:**

Choosing a model that must be used and performing it with the correct parameters are important tasks. In analytics, we have functioned unsupervised machine learning for the research. The cluster analysis is performed with two types of clusters. The task of arranging a set of objects into the same group so that they are more similar to one another than to those in other groups is known as cluster analysis or clustering.

The three types of cluster analysis achieved are:

- K-Means Clustering
- Hierarchical Clustering
- Affinity Propagation

**4.1 K-means Clustering:**

The unsupervised clustering method K-Means divides the data into k clusters, with each observation belonging to the group with the closest mean. The number of clusters we wish to divide the data into is 'K'. The term "mean" refers to the typical distance between the cluster's centre and the examples (or data points). The cluster's nucleus is referred to as the "Centroid."

Objective Function of K-Means Clustering:

$$J = \sum_{i=1}^{m} \sum_{k=1}^{K} w_{ik} \|x^i - \mu_k\|^2$$

## 4.2 Hierarchical Clustering:

Statistical methods such as hierarchical clustering analyze data by building clusters in a hierarchy. This algorithm employs two strategies, agglomerative and divisive. K can be exactly set in Hierarchical Clustering as in K-means, and n is the number of data points such that n>K. The agglomerative Hierarchical Clustering gathers data starting with n clusters and continues until K clusters are obtained. The divisive begins with a single cluster and subsequently splits into K clusters based on their commonalities.

## 4.3 Affinity Propagation Clustering:

The most popular clustering technique in data science is affinity propagation. Neither the cluster size nor the number of clusters must be provided as input. Affinity propagation also benefits from not being dependent on the luck of the first cluster centroid pick.

### 4.3.1 Responsibility Matrix(r):

Initially, create an availability matrix with all of its components set to 0. The responsibility matrix is then calculated for each cell using the following formula:

$$r(i,k) \leftarrow s(i,k) - \max_{k' \text{ such that } k' \neq k} \{a(i,k') + s(i,k')\}$$

The proceeding formula is used to fill in the elements on the diagonal:

$$a(k,k) \leftarrow \sum_{i' \text{ such that } i' \neq k} \max\{0, r(i',k)\}$$

The following equation is used to update off-diagonal elements:

$$a(i,k) \leftarrow \min\left\{0, r(k,k) + \sum_{i' \text{ such that } i' \notin \{i,k\}} \max\{0, r(i',k)\}\right\}$$

### 4.3.2 Criterion Matrix:

Simply adding the availability matrix and responsibility matrix for that location results in each cell's value in the criteria matrix.

$$c(i,k) \leftarrow r(i,k) + a(i,k)$$

**5 Methodology:**

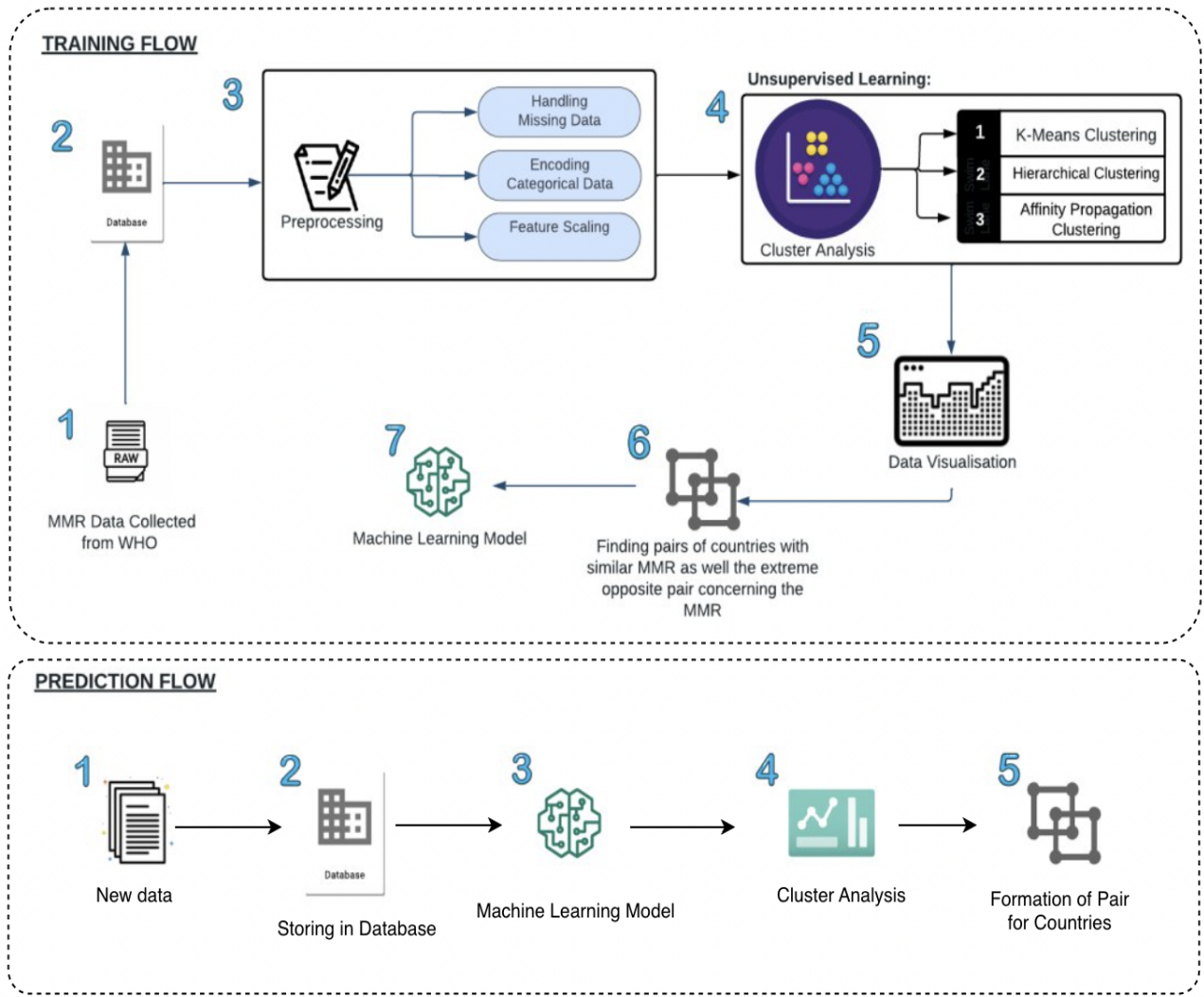

Figure 2: Methodology

**5.1 Training Flow:**
Maternal Mortality Rates from 1990 to 2015 in 185 nations are included in the data, which is gathered from the World Health Organization (WHO). The PostgreSQL database is where the data is kept. For the built-in advanced manipulation operation feature, PostgreSQL was specifically chosen.

The preprocessing steps for the data cleaning are implemented, i.e.,
- Handling Missing Data

- Encoding Categorical Data
- Feature Scaling

It is an important task to deal with missing values present in the dataset. The missing data are given equal importance as the data which is not null. The outliers present in the data are handled so that the estimations are not affected by them. All the data which is in the string format is encoded with a suitable encoding method. The data is first analysed under which category the data falls to find the exact encoding method. The feature scaling process is employed to normalise the range of features present in a dataset.

Cluster Analysis is performed to group counties which have some familiarity over the years.
Three types of Clustering performed are:
- K-Means Clustering
- Hierarchical Clustering
- Affinity Propagation Clustering

The visualisation for the cluster formation and the trend is viewed using Google Studios. The pairs of countries with similar Maternal Mortality Rates as well as the countries which are opposite concerning their Maternal Mortality are found using statistical hypothesis testing. The generated Machine Learning Model which is trained is stored for the future.

**5.2 Prediction Flow:**
The data which the model does not see is being provided for the results. The data is stored in the database and all the data preprocessing steps are implemented. The Machine Learning Model which is created in the training phase is directly used for the results. The clusters are formed for the new data with the help of the model. Countries with equivalent Maternal Mortality Rate (MMR) are grouped. The countries with similar Maternal Mortality Rate trends over the period are mapped as pairs and the pairs with the exact opposite trend in the Maternal Mortality Rate are also mapped as pairs.

**6 Tech Stack:**
**6.1 Streamlit:**
A Python-based toolkit called Streamlit makes it simple for data scientists to produce open-source Machine-learning applications. An uncomplicated and user-friendly interface makes it simple to read and interact with a saved model. So streamlit is used here for the result display when the new data is provided.

**6.2 Docker:**
The open-source programme Docker is employed during the application's testing, development, delivery, and operation. Docker made the process of building the application more efficient. Docker usage was required because there are numerous microservices.

**6.3 Google Data Studio:**
Google Data Studio is a visualisation and reports creation tool. It allows users to use the enhanced features, Google Data Studio is used for the development of impactful visualisations that pleasingly depict the findings.

**6.4 Machine Learning Libraries:**
**6.4.1 Pandas:**
Pandas is built on top of the Python programming language. Pandas are a powerful tool used for data analysis and manipulation. Here pandas are used to complete the data manipulation task faster and more effectively.

**6.4.2 NumPy:**
NumPy is an open-source library for Python programming language. It supports huge arrays which are in multi-dimension form.

**6.4.3 Scikit-learn:**
The Scikit-learn is a free software ML library which effectively works to perform supervised learning tasks as well as unsupervised learning tasks. In this study, we have performed clustering which falls under unsupervised learning with the scikit-learn library.

**6.4.4 Matplotlib:**
Matplotlib is a Python library used for visualising the data. It provides an object-oriented API.

**6.4.5 TSNE:**
The t-Distributed Stochastic Neighbor Embedding (t-SNE) is a tool in the Python programming language used to visualise high-dimensional data.

**6.4.6 Seaborn:**

Seaborn is a Python data visualisation library. It gives access to a high-level interface for informative and attractive visualisation of data.

**7 RESULTS AND ANALYSIS:**

The Clustering process followed by pairing the geographical location is implemented and the results are

**K-Means Clustering Visualisation:**

The data points are grouped with the help of the K-Means clustering method. The centroid is chosen with the implementation of two methods namely the Elbow method and the Silhouette method.

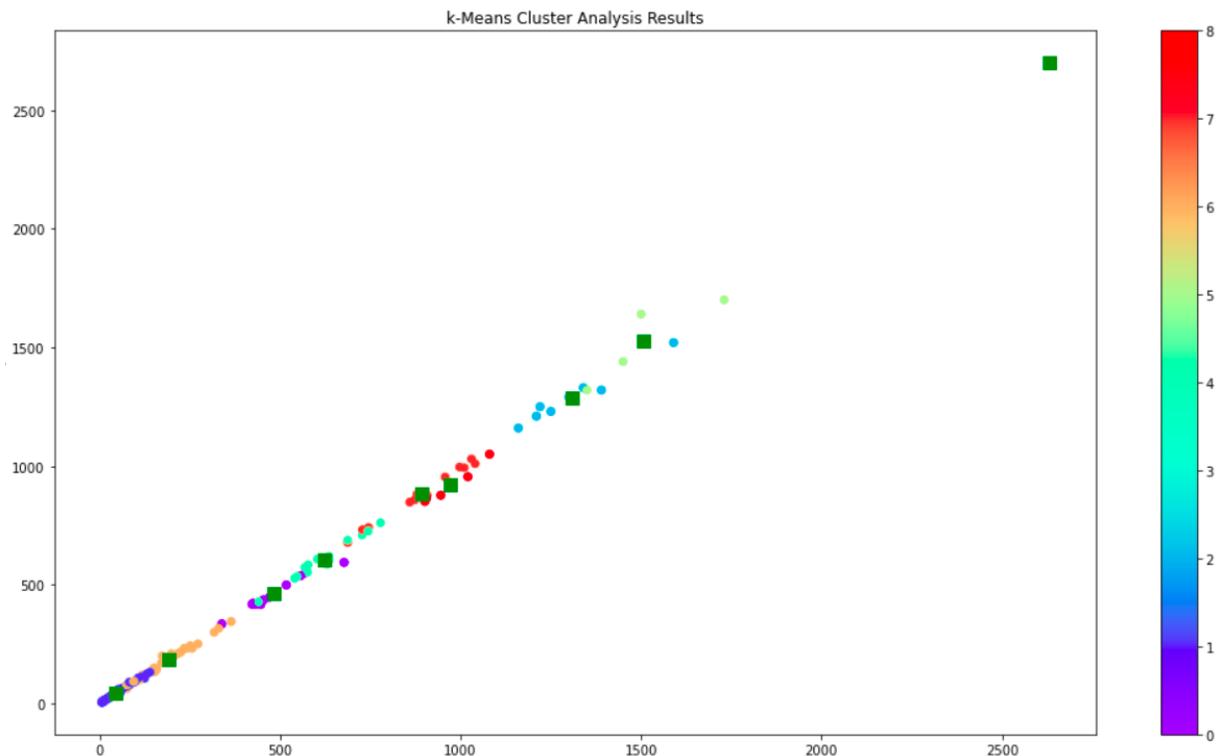

Figure 3: K-Means Clustering Result Visualization

**Hierarchical Clustering Visualisation:**

The clusters are formed using the Hierarchical Clustering method and visualisation is effectively made.

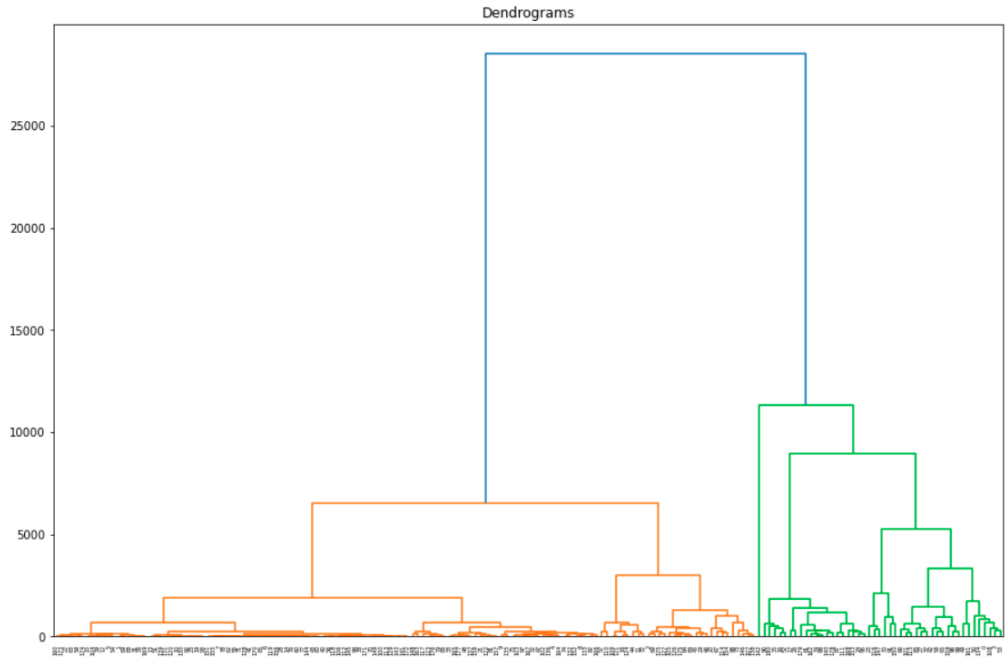

Figure 4: Hierarchical Clustering Result Visualization

| # | Country-1 | Country-2 |
|---|---|---|
| 1 | Albania | Algeria |
| 2 | Albania | Cyprus |
| 3 | Albania | Honduras |
| 4 | Albania | Iran |
| 5 | Albania | Iraq |
| 6 | Albania | Ireland |
| 7 | Albania | Jamaica |
| 8 | Albania | Jordan |
| 9 | Albania | Moldova |
| 10 | Albania | Montenegro |
| 11 | Albania | Philippines |
| 12 | Albania | Syria |
| 13 | Albania | Ukraine |
| 14 | Albania | Vietnam |
| 15 | Afghanistan | Democratic Republic of Congo |
| 16 | Afghanistan | Ethiopia |
| 17 | Afghanistan | Gambia |
| 18 | Afghanistan | Guinea |
| 19 | Afghanistan | Malawi |
| 20 | Afghanistan | Tanzania |
| 21 | Benin | Comoros |
| 22 | Benin | Laos |
| 23 | Benin | Madagascar |
| 24 | Benin | Togo |
| 25 | Benin | Yemen |

Figure 5: Countries with extremely opposite concerning the MMR

| # | Country - 1 | Country - 2 | | # | Country - 1 | Country - 2 |
|---|---|---|---|---|---|---|
| 1 | Albania | North Korea | | 40 | Albania | United Arab Emirates |
| 2 | Albania | North Macedonia | | 41 | Albania | United Kingdom |
| 3 | Albania | Norway | | 42 | Albania | United States |
| 4 | Albania | Oman | | 43 | Albania | Uruguay |
| 5 | Albania | Pakistan | | 44 | Albania | Uzbekistan |
| 6 | Albania | Palestine | | 45 | Albania | Vanuatu |
| 7 | Albania | Panama | | 46 | Albania | Venezuela |
| 8 | Albania | Papa New Guinea | | 47 | Afghanistan | Angola |
| 9 | Albania | Paraguay | | 48 | Afghanistan | Burundi |
| 10 | Albania | Peru | | 49 | Afghanistan | Cameroon |
| 11 | Albania | Poland | | 50 | Afghanistan | Chad |
| 12 | Albania | Portugal | | 51 | Afghanistan | Cote d'Ivoire |
| 13 | Albania | Puerto Rico | | 52 | Afghanistan | Equatorial Guinea |
| 14 | Albania | Qatar | | 53 | Afghanistan | Eritrea |
| 15 | Albania | Romania | | 54 | Afghanistan | Guinea-Bissau |
| 16 | Albania | Russia | | 55 | Afghanistan | Kenya |
| 17 | Albania | Saint Lucia | | 56 | Afghanistan | Liberia |
| 18 | Albania | Saint Vincent | | 57 | Afghanistan | Mali |
| 19 | Albania | Samoa | | 58 | Afghanistan | Mauritania |
| 20 | Albania | So Tome and Principe | | 59 | Afghanistan | Mozambique |
| 21 | Albania | Saudi Arabia | | 60 | Afghanistan | Niger |
| 22 | Albania | Serbia | | 61 | Afghanistan | Nigeria |
| 23 | Albania | Singapore | | 62 | Afghanistan | Rwanda |
| 24 | Albania | Slovakia | | 63 | Afghanistan | Somalia |
| 25 | Albania | Solomon Islands | | 64 | Afghanistan | South Sudan |
| 26 | Albania | South Africa | | 65 | Afghanistan | Timor |
| 27 | Albania | South Korea | | 66 | Benin | Bhutan |
| 28 | Albania | Spain | | 67 | Benin | Burkina Faso |
| 29 | Albania | Sri Lanka | | 68 | Benin | Cambodia |
| 30 | Albania | Suriname | | 69 | Benin | Congo |
| 31 | Albania | Sweden | | 70 | Benin | Eswatini |
| 32 | Albania | Switzerland | | 71 | Benin | Ghana |
| 33 | Albania | Tajikistan | | 72 | Benin | Haiti |
| 34 | Albania | Thailand | | 73 | Benin | Lesotho |
| 35 | Albania | Tonga | | 74 | Benin | Nepal |
| 36 | Albania | Trinidad and Tobago | | 75 | Benin | Senegal |
| 37 | Albania | Tunisia | | 76 | Benin | Sudan |
| 38 | Albania | Turkey | | 77 | Benin | Uganda |
| 39 | Albania | Turkmenistan | | 78 | Benin | Zambia |
| | | | | 79 | Benin | Zimbabwe |

Figure 6: Countries with similar concerning the MMR

## 8 Conclusion:

The direct principle behind this conducted research is the process for the pregnant women, members related to her, medics and particularly the government to create a more satisfactory than the existing surroundings to protect against unnecessary death. The time series forecasting can be executed and the same cluster analysis can be performed to find what the future is more likely to have for maternal mortality. The most prominent benefit is that the same research can be continued just by adding more additional country data as well the more MMR. The project implementation was feasible only if the person functioning in this research has a satisfactory amount of familiarity with Machine Learning because the methodology incorporates data cleaning, data visualization, and the use of unsupervised ML. Since this research includes advanced approaches, the study can be further developed and taken to the next level by using more methods and adding to them. This research will consistently benefit society because of the sensitive issue of maternal mortality. This research is feasible without any usage of money because all the methods used are open source only the collection of data will consume a bit of time.